\documentclass{article}

\usepackage{arxiv}
\usepackage{xcolor}
\usepackage[utf8]{inputenc} 
\usepackage[T1]{fontenc}    
\usepackage{hyperref}       
\usepackage{url}            
\usepackage{booktabs}       
\usepackage{amsfonts}       
\usepackage{nicefrac}       
\usepackage{microtype}      
\usepackage{lipsum}		
\usepackage{graphicx}
\usepackage{natbib}
\usepackage{doi}
\usepackage[ruled]{algorithm2e}
\usepackage{algpseudocode}
\usepackage{fullpage}
\usepackage{subcaption}
\usepackage{times}
\usepackage{fancyhdr,graphicx,amsmath,amssymb}
\usepackage{wrapfig,lipsum,booktabs}
\include{pythonlisting}

\title{Faster Segment Anything:\\Towards Lightweight SAM for Mobile Applications} 

\title{Faster Segment Anything: Towards Lightweight SAM for Mobile Applications} 

\author{
Chaoning Zhang\thanks{You are welcome to contact the authors through chaoningzhang1990@gmail.com} \\
	Kyung Hee University \\
\And
Dongshen Han \\
	Kyung Hee University\\
\And
Yu Qiao \\
	Kyung Hee University\\
\And
Jung Uk Kim \\
	Kyung Hee University\\
  \And
Sung-Ho Bae \\
	Kyung Hee University\\
   \And
Seungkyu Lee \\
	Kyung Hee University\\
 \And
Choong Seon Hong\\
Kyung Hee University\\
}

\begin{document}
\maketitle

\begin{abstract}
Segment Anything Model (SAM) has attracted significant attention due to its impressive zero-shot transfer performance and high versatility for numerous vision applications (like image editing with fine-grained control). Many of such applications need to be run on resource-constraint edge devices, like mobile phones. In this work, we aim to make SAM mobile-friendly by replacing the heavyweight image encoder with a lightweight one. A naive way to train such a new SAM as in the original SAM paper leads to unsatisfactory performance, especially when limited training sources are available. We find that this is mainly caused by the coupled optimization of the image encoder and mask decoder, motivated by which we propose decoupled distillation. Concretely, we distill the knowledge from the heavy image encoder (ViT-H in the original SAM) to a lightweight image encoder, which can be automatically compatible with the mask decoder in the original SAM. The training can be completed on a single GPU within less than one day, and the resulting lightweight SAM is termed MobileSAM which is more than 60 times smaller yet performs on par with the original SAM. For inference speed, With a single GPU, MobileSAM runs around 10ms per image: 8ms on the image encoder and 4ms on the mask decoder. With superior performance, our MobileSAM is around 5 times faster than the concurrent FastSAM and 7 times smaller, making it more suitable for mobile applications. Moreover, we show that MobileSAM can run relatively smoothly on CPU. 
The code for our project is provided at 
\href{https://github.com/ChaoningZhang/MobileSAM}{\textcolor{red}{MobileSAM}}), with a demo showing that MobileSAM can run relatively smoothly on CPU. 

\end{abstract}

\section{Introduction}
\begin{figure}[!htbp]
     \centering
     \begin{minipage}[b]{\textwidth}
         \centering
         \includegraphics[width=\textwidth]{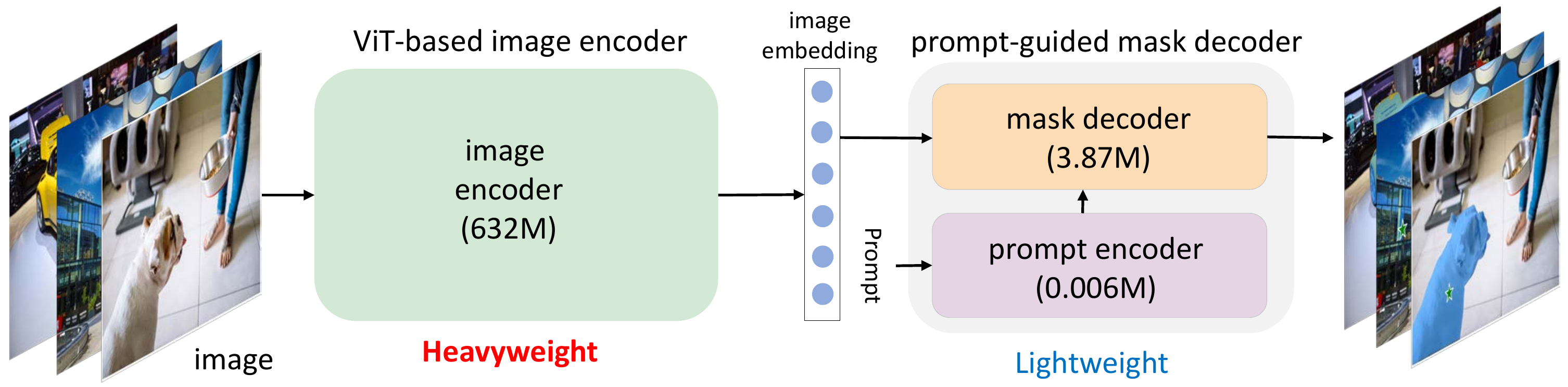}
     \end{minipage}
        \caption{The overview of Segment Anything Model.  
        }
    \label{fig:SAM}
\end{figure}

ChatGPT~\cite{zhang2023ChatGPT} has revolutionized the NLP field, marking a breakthrough in generative AI (AIGC, a.k.a Artificial intelligence generated content)~\cite{zhang2023complete}. What has made this possible lies in GPT-series models~\cite{brown2020language,radford2018improving,radford2019language}, which are foundation models~\cite{bommasani2021opportunities} trained on web-scale text datasets. Following the success of foundation models in NLP, multiple works~\cite{he2020momentum,qiao2023mp,zhang2022how} have learned an image encoder together with a text encoder via contrastive learning~\cite{he2020momentum,zhang2022dual}. Very recently, Meta Research team has released the "Segment Anything" project~\cite {kirillov2023segment}, where a prompt-guided vision foundation termed SAM has been proposed and is believed to be a GPT moment for vision. SAM consists of two components: ViT-based image encoder and prompt-guided mask decoder, which work in sequence (see Figure~\ref{fig:SAM}).

Since its advent, SAM has attracted significant attention for multiple reasons. First, it is the first to show that vision can follow NLP to pursue a path that combines foundation model with prompt engineering. Second, it is the first to perform label-free segmentation, a fundamental vision task that is in parallel to label prediction~\cite{zhang2023samsurvey}. Moreover, this fundamental task makes SAM compatible with other models to realize advanced vision applications, like text-guided segmentation and image editing with fine-grained control. Many of such use cases, however, need to be run on resource-constrained edge-devices, like mobile apps. As shown in the official~\href{https://segment-anything.com/}{demo}, with a processed image embedding, the SAM can work
on resource-constrained devices because the mask decoder is lightweight. What makes the SAM pipeline computation heavy lies in the huge image encoder. In this work, we investigate how to obtain a lightweight SAM suitable for resource-constrained mobile devices, which is therefore termed MobileSAM. 

\begin{wraptable}{r}{11cm}
\centering
\caption{Parameters SAM with different image encoders.}
\label{tab:model_param}
\begin{tabular}{ccccc}
\toprule
Parameters & SAM (ViT-H) & SAM (ViT-L) & SAM (ViT-B)  \\ 
\midrule
ViT-based encoder & 632M & 307M & 86M \\
prompt-guided encoder & 0.006M & 0.006M & 0.006M \\
\bottomrule
\end{tabular}
\end{wraptable}

Given that the default image encoder in the SAM is based on ViT-H, a straightforward way to obtain MobileSAM is to follow the official pipeline in~\cite{kirillov2023segment}
to retrain a new SAM with a smaller image encoder like replacing the ViT-H with a smaller ViT-L or even smaller ViT-B. The parameters of SAM with different variants of image encoder are summarized in Table~\ref{tab:model_param}. As stated in~\cite{kirillov2023segment}, training a new SAM with ViT-L or ViT-B as the image encoder requires 128 GPUs for multiple days. Such resource-intensive retraining can be a non-trivial burden to reproduce or improve their results. This optimization difficulty mainly comes from the coupled optimization of the image encoder and mask decoder. Motivated by this understanding, we propose to decouple the optimization of the image encoder and mask decoder. Concretely, we first distill the knowledge from the default image encoder ViT-H to a tiny ViT. After that, we can finetune the mask decoder in the original SAM to better align with the distilled image encoder. It is worth highlighting that the alignment optimization is optional because the fact that the lightweight image encoder is distilled from the default image encoder guarantees its inherent alignment with the default mask decoder.

By turning the problem of seeking a new SAM pipeline into a decoupled distillation, our approach has the advantage of being simple, and effective, while being reproducible at a low cost (on a single GPU with less than a day). The resulting MobileSAM reduces the encoder parameters by 100 times and total parameters by 60 times yet. Surprisingly, such a lightweight MobileSAM performs on par with the original heavyweight SAMs, which constitutes a significant step for pushing SAM for mobile applications. For the inference with MobileSAM, a single image takes runs only around 10ms: 8ms on the image encoder and 4ms on the mask decoder. It is worth highlighting that our MobileSAM is around 5 times faster and 7 times smaller than
the concurrent FastSAM~\cite{zhao2023fast}, while achieving superior performance.

\section{Related work} \label{sec:related}
\paragraph{SAM: generalization and versatility.} Since its advent in early April of this year, numerous projects and papers have emerged to investigate SAM from different perspectives. Given that SAM claims to segment anything, a line of works has reported its performance in real-world situations, including medical images~\cite{ma2023segment,zhang2023input}, camouflaged objects~\cite{tang2023can}, and transparent objects~\cite{han2023segment}. The findings consistently show that SAM works well in general setups, but not in the above challenging setups. Another significant research direction has focused on enhancing SAM to improve its practicality. Attack-SAM~\cite{zhang2023attacksam} has shown that the output masks of SAM can be easily manipulated by adversarial attacks through maliciously generated adversarial perturbations. Another work ~\cite{qiao2023robustness} further conducts a comprehensive robustness evaluation of SAM, ranging from style transfer and common corruptions to local occlusion and adversarial perturbation. It is found in~\cite{qiao2023robustness} SAM has high robustness but not for adversarial perturbation, which aligns well with the finding in~\cite{zhang2023attacksam}. Another line of work focuses on demonstrating the versatility of SAM. Grounded SAM~\cite{GroundedSegmentAnything2023} is the pioneering work to combine Grounding DINO~\cite{liu2023grounding} with SAM for segmenting anything with text inputs. Specifically, it relies on Grounding DINO to generate a bounding box from text and then the generated box can be used as a prompt to segment the mask. SAM predicts masks without labels and multiple works~\cite{chen2023semantic,park2023segment} combine SAM with other models such as CLIP~\cite{radford2021learning} to semantically segment anything. Beyond object segmentation, multiple works have also shown its versatility in other fields, including image editing~\cite{rombach2022high}, as well as inpainting tasks~\cite{yu2023inpaint}, object tracking within videos~\cite{yang2023track,z-x-yang_2023}. Beyond 2D vision, the investigation of SAM has also been extended to 3D object reconstruction~\cite{shen2023anything,kang2022any}, demonstrating its capabilities in assisting 3D model generation from a single image. For a complete survey of SAM, the readers are suggested to refer to~\cite{zhang2023samsurvey}.

\paragraph{ViT: lightweight and efficient.} Early mobile vision applications have been mainly powered by lightweight CNNs, such as MobileNet~\cite{howard2017mobilenets} and its improved varinats~\cite{sandler2018mobilenetv2,howard2019searching}. The core idea of MobileNet lies in separating a normal convolution block into depth-wise convolution and point-wise convolution, which significantly reduces the mode parameters and computation time. Since the advent of ViT~\cite{dosovitskiy2021an}, numerous works have attempted to make it lightweight and efficient. Complementary to ViT-Huge (ViT-H), ViT-Large (ViT-L), ViT-Base (ViT-B) in the original ViT paper~\cite{dosovitskiy2021an}, smaller ViTs are introduced in~\cite{touvron2020training} and are denoted as Deit-Small (Deit-S) and Deit-Tiny (Deit-T) ViT-Small and ViT-Tiny. MobileViT~\cite{mehta2021mobilevit} is a pioneering work to combine ViT with standard convolutions for improving its performance, which outperforms MobileNet v2~\cite{sandler2018mobilenetv2}. The main motivation is to exploit the local representation capability of CNN, and this practice is followed by multiple follow-up works which aim to enhance the model speed, including EfficientFormer~\cite{li2022efficientformer}, EfficientViT~\cite{liu2023efficientvit}, Next-ViT~\cite{li2022nextvit} and Tiny-ViT~\cite{wu2022tinyvit}. The recent progress in lightweight and faster ViT is complementary to our proposed decoupled distillation towards making the next-generation SAM suitable for resource-constrained mobile devices. 

\section{Mobile-Friendly SAM} 

\subsection{Background and Project Goal}

\paragraph{Background on SAM.} Here, we first summarize the structure of SAM and how it works. SAM consists of a ViT-based image encoder and a prompt-guided mask decoder. The image encoder takes the image as the input and generates an embedding, which is then fed to the mask decoder. The mask decoder generates a mask to cut out any object from the background based on a prompt like a point (or box). Moreover, SAM allows generating multiple masks for the same prompt for addressing the ambiguity issue, which provides valuable flexibility. Considering this, this work maintains the pipeline of SAM to first adopt a ViT-based encoder to generate image embedding and then to adopt a prompt-guided decoder to generate the desired mask. This pipeline is optimally designed for the ``segment anything", which can be used for the downstream task of  ``segment everything" (see Sec.~\ref{sec:comparison} for more discussion).

\paragraph{Project goal.} The goal of this project is to generate a mobile-friendly SAM (MobileSAM) that achieves satisfactory performance in a lightweight manner and is much faster than the original SAM. The prompt-guided mask decoder in the original SAM has less than 4M parameters and is thus considered lightweight. Given an image embedding processed by the encoder, as shown in their public demo, SAM can work in resource-constrained devices since the mask decoder is lightweight. However, the default image encoder in the original SAM is based on ViT-H with more than 600M parameters, which is very heavyweight and makes the whole SAM pipeline incompatible with mobile devices. Therefore, the key to obtaining a mobile-friendly SAM lies in replacing the heavyweight image encoder with a lightweight one, which also automatically keeps all its functions and characteristics of the original SAM. In the following, we elaborate on our proposed method for achieving this project goal. 

\subsection{Proposed Method}
\begin{figure}[!htbp]
     \centering
     \begin{minipage}[b]{\textwidth}
         \centering
         \includegraphics[width=0.50\textwidth]{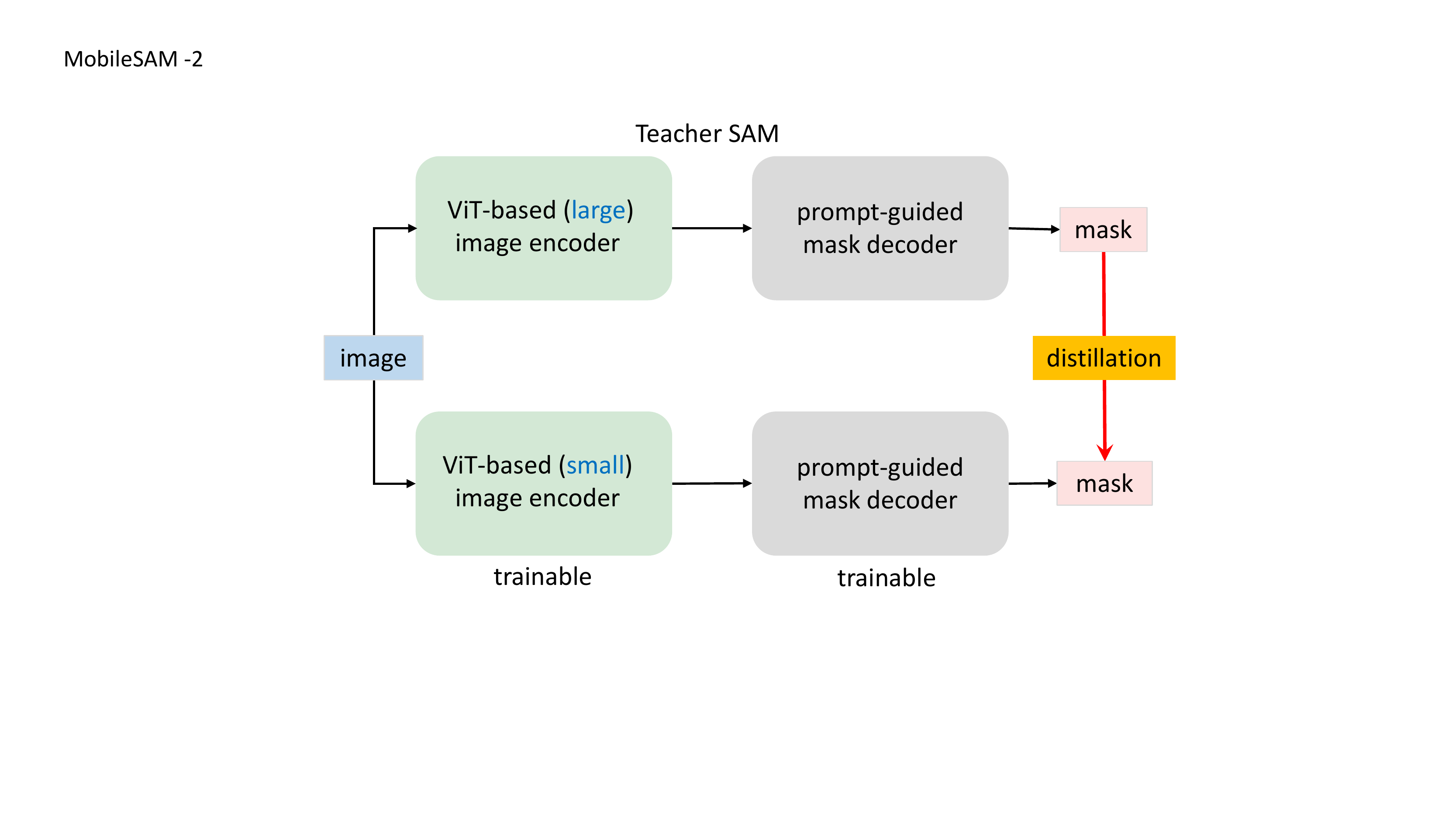}
        \includegraphics[width=0.49\textwidth]{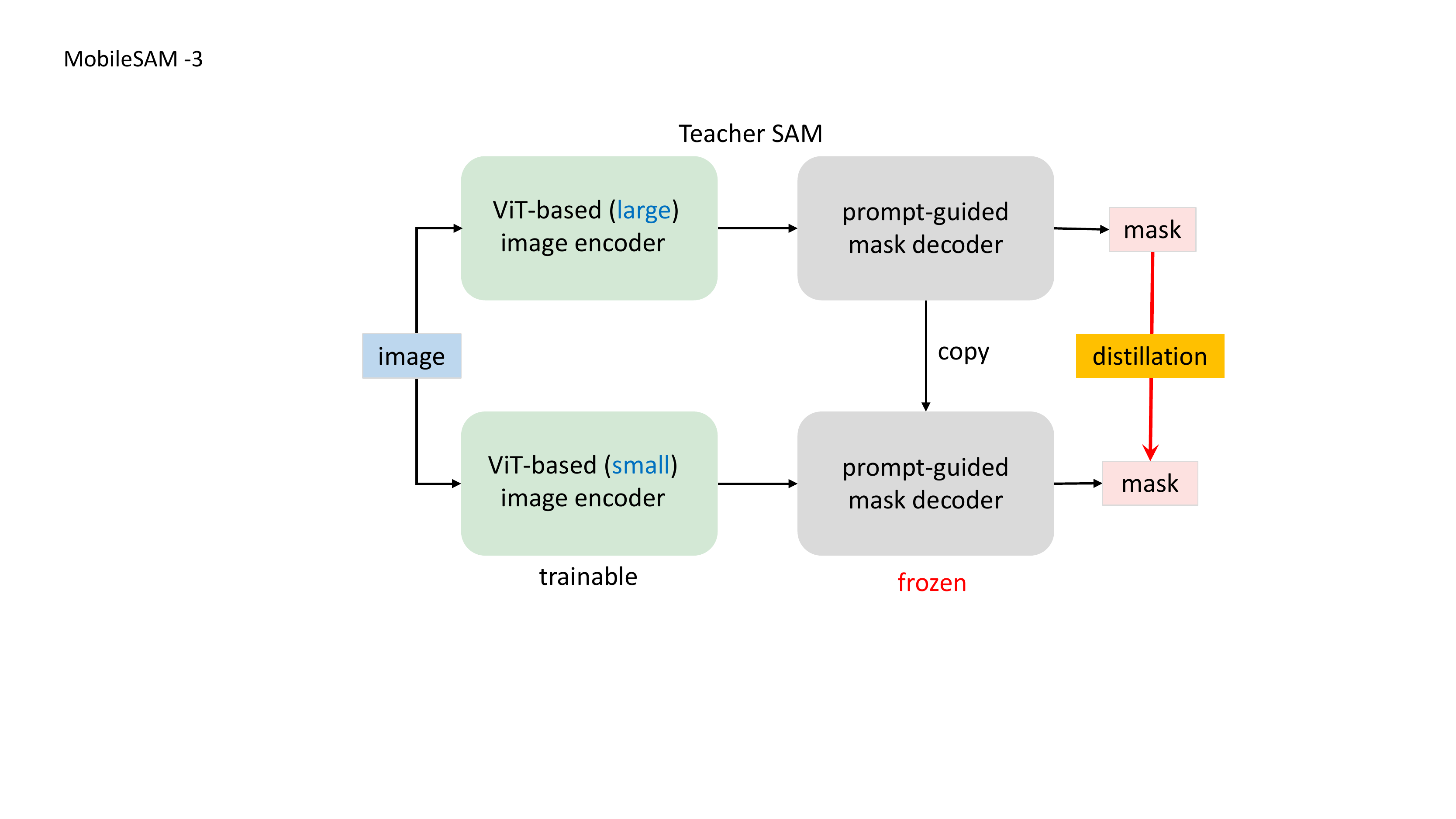}
     \end{minipage}
        \caption{Coupled knowledge distillation of SAM. The left subfigure denotes the fully-coupled distillation, while the right one represents the semi-coupled distillation.
        }
    \label{fig:coupled_distillation}
\end{figure}

\paragraph{Coupled distillation.} A straightforward way to realize our project goal is to follow the official pipeline in~\cite{kirillov2023segment} to retrain a new SAM with a smaller image encoder. As stated in~\cite{kirillov2023segment}, training a SAM with ViT-H image encoder requires takes 68 hours on 256 A100 GPUs. Replacing the ViT-H with ViT-L or ViT-B reduces the required GPUs to 128, which is still a non-trivial burden for many researchers in the community to reproduce or improve their results. Following their approach, we can further adopt an even smaller image encoder and retrain a new SAM with their provided segmentation dataset which is 11-T. Note that the masks in the provided dataset are given by the pretrained SAM (with the ViT image encoder). In essence, this retraining process is \textit{knowledge distillation}~\cite{hinton2015distilling}, which transfers the knowledge from a ViT-H-based SAM to a SAM with a smaller image encoder (see Figure~\ref{fig:coupled_distillation} left). 

\begin{figure}[!htbp]
     \centering
     \begin{minipage}[b]{\textwidth}
         \centering
         \includegraphics[width=\textwidth]{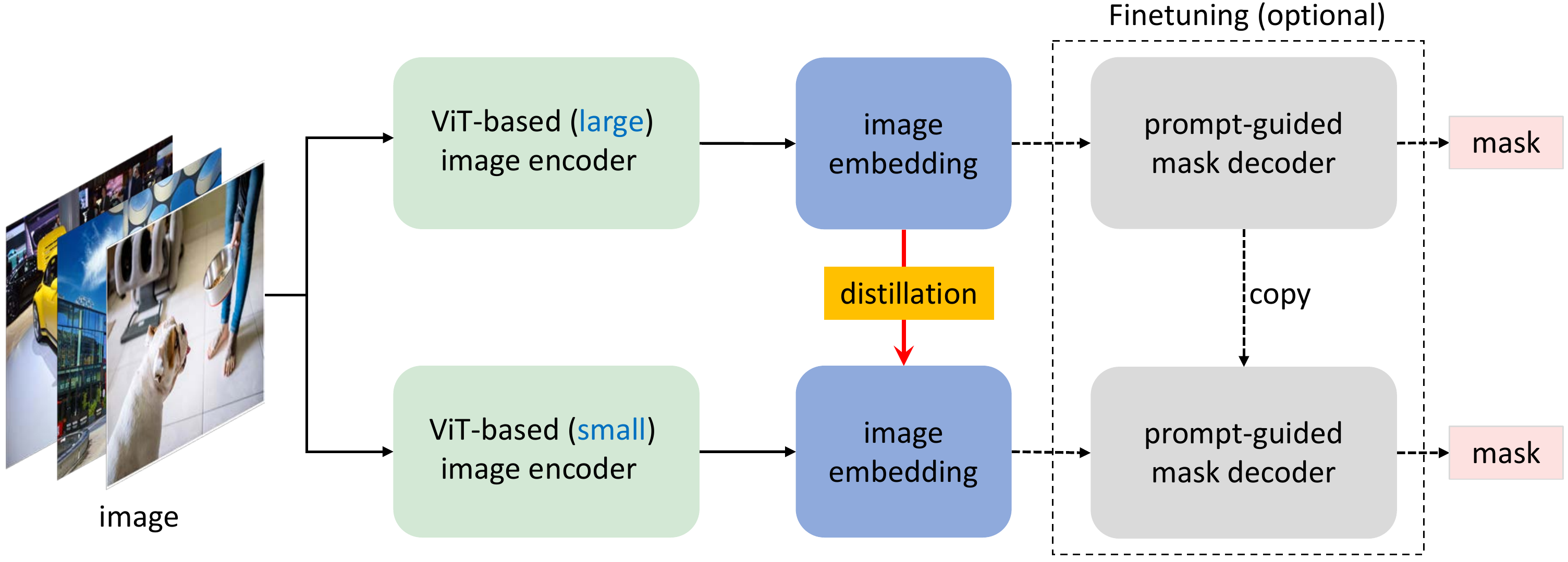}
     \end{minipage}
        \caption{Decoupled distillation for SAM.  
        }
    \label{fig:decoupled_distillation}
\end{figure}

\paragraph{From semi-coupled to decoupled distillation.} When performing a KD from the original SAM to that with a smaller image encoder, the difficulty mainly lies in a coupled optimization of the image encoder and combined decoder. Intuitively, the optimization of the image encoder depends on the quality of the image decoder, and vice versa. When the two modules in the SAM are both in a bad state, it is more challenging to train them both to a good state. Inspired by the divide-and-conquer algorithm~\cite{zhang2022decoupled}, we propose to divide the KD task into two sub-tasks: image encoder distillation and mask decoder finetuning. Concretely, we first perform the KD on the image encoder by transferring the knowledge from ViT-H to a smaller encoder. Since the mask decoder in the original SAM is already lightweight, we plan to keep its architecture. This brings a benefit of a readily used combined decoder for finetuning instead of training it from scratch. To alleviate the optimization issue of coupled distillation, a straightforward way is to optimize the image encoder with a copied and frozen mask decoder (see Figure~\ref{fig:coupled_distillation} right). The freezing operation can help prevent the quality of the mask decoder from being deteriorated by a poor image encoder. We call this distillation semi-coupled because the optimization of the image encoder is still not fully decoupled from the mask decoder. Empirically, we find that this optimization is still challenging because the choice of a prompt is random, which makes the mask decoder variable and thus increases the optimization difficulty. Therefore, we propose to distill the small image encoder directly from the ViT-H in the original SAM without resorting to the combined decoder, which is termed decoupled distillation (see Figure~\ref{fig:decoupled_distillation}). Another advantage of performing distillation on the image embedding is that we can adopt a simple MSE loss instead of using a combination of focal loss~\cite{lin2017focal} and dice loss~\cite{milletari2016v} for making the mask prediction as in~\cite{kirillov2023segment}. 

\paragraph{On the necessity of mask decoder finetuning.} Unlike the semi-coupled distillation, the above decoupled distillation yields a lightweight image encoder that might not align well with the original frozen mask decoder. Empirically, we find that this is not true because the generated image encoding from the student image encoder can be sufficiently close to that of the original teacher encoder, which renders finetuning on the combined decoder in the second stage optional. It is expected that finetuning the mask decoder on the frozen lightweight image encoder or jointly finetuning them together might further improve the performance. 

\paragraph{Preliminary evaluation.} Here, we conduct a preliminary investigation to compare coupled distillation and decoupled distillation. Here, for performance evaluation, we compute the mIoU between the two masks generated by the teacher SAM and student SAM on the same prompt point. Intuitively, a higher mIoU indicates a higher mask prediction performance by assuming that the mask generated by ViT-H is ground-truth. For the coupled distillation, we adopt the SAM with ViT-B provided in the original SAM~\cite{kirillov2023segment}. It was trained on SA-1B (11M images) on 128 GPUs (1 sample per GPU) for 180k iterations. By contrast, in our decoupled distillation setup, we train the model on 2 GPUs (two samples per GPU to save computation resources) on 0.1\% samples of SA-1B dataset (11k) images for 55k iterations. Overall, decoupled distillation takes less than 1\% of the computation resources than coupled distillation, while achieving a superior performance of mIoU of 0.75 vs 0.72 for the coupled sit (averaged on 200 samples). Since ViT-B is still a non-trivial burden for mobile devices, therefore in the following we experiment with a TinyViT (with 5M parameters)~\cite{wu2022tinyvit} based on our proposed decoupled distillation.

\begin{table}[!htp]
\centering
\caption{Comparison of coupled distillation and decoupled distillation fro SAM with ViT-B as the image encoder. Decoupled distillation performs better and requires less than 1\% computation resources than coupled distillation.}
\label{tab:weather_miou}
\begin{tabular}{cccccc}
\toprule
/ & SAM (coupled distillation)  & SAM (decoupled distillation)  \\ 
\midrule
MIoU  & 0.72 & 0.75 \\
\midrule
Training GPUs & 128  & 2 \\
Batch size  & 128 & 4 \\
Iterations & 180k & 55k \\
Training Data & 11M & 11K \\
\bottomrule
\end{tabular}
\end{table}

\section{Experiments}

\subsection{Experimental Setup}
\begin{wraptable}{r}{10cm}
\centering
\caption{Comparison of the parameters and speed for the image encoder in original SAM and MobileSAM. The inference speed is measured on a single GPU.}
\label{tab:mobilesam_param}
\begin{tabular}{ccccc}
\toprule
 & Original SAM & MobileSAM &  \\ 
\midrule
Parameters & 632M & 5.78M \\
Speed & ~452ms & ~8ms \\ 
\bottomrule
\end{tabular}
\end{wraptable}
\paragraph{Lightweight Image Encoder.} The goal of our project is to obtain an efficient SAM by replacing the default ViT-H with a lightweight image encoder for mobile devices. As a ViT-based backbone, ViT-Tiny has similar parameters as Deit-Tiny but performs better. For example, on ImageNet-1K, Deit-Yiny achieves an accuracy of 72.2\%, while ViT-Tiny achieves 79.1\%. Therefore, we adopt ViT-Tiny for the proof of concept to demonstrate the effectiveness of our proposed decoupled distillation for training a lightweight MobileSAM that can be much faster than the original SAM. The adopted lightweight image encoder consists of four stages which gradually reduce the resolution. The first stage is constructed by convolution blocks with inverted residuals~\cite{sandler2018mobilenetv2}, while the remaining three stages consist of transformer blocks. At the beginning of the model, there are 2 convolutions blocks with a stride of 2 for downsampling the resolution. The downsampling operation between different stages is processed by convolution blocks with the stride of 2. Different from~\cite{wu2022tinyvit}, we set the stride of 2 in the last downsampling convolution to 1 for making the final resolution match that of the ViT-H image encoder of the original SAM. The parameters inference speed of MobileSAM are summarized in Table~\ref{tab:mobilesam_param}. Note that other efficient image encoders discussed in Section~\ref{sec:related} can also be adopted as the image encoder.

\paragraph{Training and evaluation details.} For the decoupled KD on the image encoder, we train the lightweight encoder with 1\% of the SA-1B dataset~\cite{kirillov2023segment} for 8 epochs on a single GPU. We observe that more computation is spent on the forward process on the teacher image encoder considering that it is significantly more heavy than our adopted student image encoder (see above). To make the distillation faster, we follow the practice in~\cite{wu2022tinyvit} to save the image embeddings beforehand so that we only need to run the forward process once. With a single GPU, we can obtain our MobileSAM in less than a day. Training our MobileSAM with more GPUs for a longer time is expected to yield better performance. The initial investigation of performing mask decoder finetuning further improves the performance of MobileSAM, however, we omit this step in this version of our paper for simplicity. For quantitative evaluation of the distilled SAM, we compute the mIoU between the masks predicted by the original SAM and our MobileSAM.

\begin{figure*}[!htbp]
     \centering
    \begin{minipage}[t]{0.9\textwidth}
         \includegraphics[width=\textwidth]{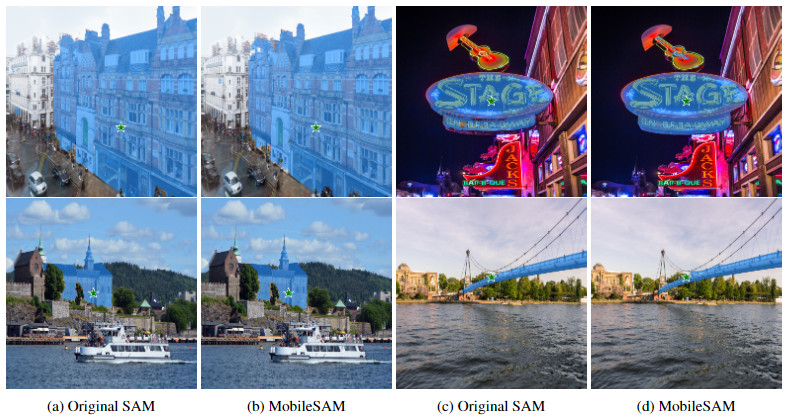}
     \end{minipage}
        \caption{Mask prediction with a single point as the prompt.}
    \label{fig:point}
\end{figure*}

\subsection{MobileSAM performs on par with the orignal SAM}
\begin{figure*}[!htbp]
     \centering
    \begin{minipage}[t]{0.9\textwidth}
         \includegraphics[width=\textwidth]{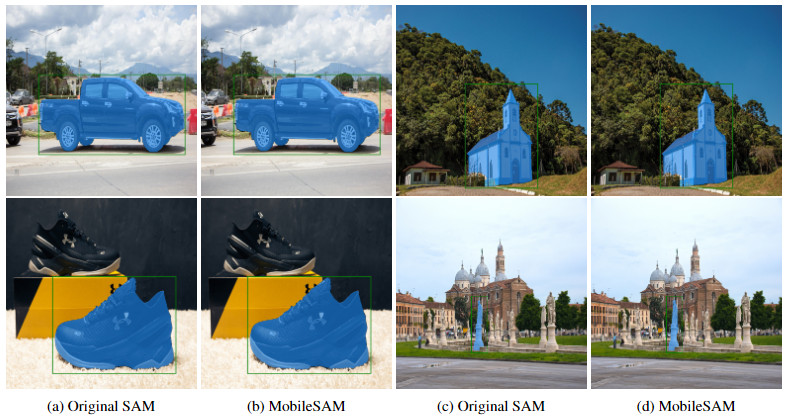}
     \end{minipage}

        \caption{Mask prediction with a box as the prompt.}
    \label{fig:box}
\end{figure*}

For the main results, we report the predicted masks with two types of prompts: point and box. We do not report the results with text prompt because the official github project of SAM does not provide pretrained models for text-guided mask decoder. The results with point as the prompt are shown in Figure~\ref{fig:point}, and those with box as the prompt are shown in Figure~\ref{fig:box}. We observe that MobileSAM makes a satisfactory mask prediction similar to that of the original SAM.

\begin{wraptable}{r}{6cm}
\centering
\caption{Ablation study on the influence of training computation on the MobileSAM performance.}
\label{tab:ablation}
\begin{tabular}{cccccc}
\toprule
 batch size  & epochs & Iterations & mIoU \\ 
\midrule
4  & 2 & 50k & 0.7057\\
8  & 4 & 50k & 0.7286\\
8  & 8 & 100k & 0.7447\\
\bottomrule
\end{tabular}
\end{wraptable}
\paragraph{Ablation study.} Here, we conduct an ablation study on the influence of the training computation on the performance of SAM. The results in Table~\ref{tab:ablation} show that, under the same number of iterations, increasing the batch size increases the model performance. Moreover, under the batch size, the performance also benefits from more update iterations by increasing the training epochs. Note that all the experiments are conducted on a single GPU. We expect that increasing the number of GPUs for allowing a larger batch size or further increasing the iterations can further improve the performance.

\subsection{MobileSAM outperforms FastSAM} \label{sec:comparison}

\begin{wraptable}{r}{6cm}
\centering
\caption{Comparison between segment anything and segment everything.}
\label{tab:model_mode}
\begin{tabular}{ccccc}
\toprule
 & anything & everything  \\ 
\midrule
\# of objects & 1 & N \\
prompt-aware & yes & no \\
\bottomrule
\end{tabular}
\end{wraptable}
\paragraph{Segment anything \textit{v.s.} segment everything .} Note that the title of the original SAM paper~\cite{kirillov2023segment} is ``segment anything" instead of ``segment everything". As highlighted in~\cite{kirillov2023segment}, SAM performs the task of promptable segmentation which ``returns a valid segmentation mask given any segmentation prompt" (quote from~\cite{kirillov2023segment}). The role of the prompt is to specify what to segment in the image. In theory, any object can be segmented as long as the prompt is set properly, therefore, it is called ``segment anything". By contrast, ``segment everything" is in essence object proposal generation~\cite{kirillov2023segment}, for which the prompt is not necessary. In~\cite{kirillov2023segment}, ``segment everything" (object proposal generation) is chosen as one of the downstream tasks for demonstrating its zero-shot transfer performance. To summarize, ``segment anything" solves the foundation task of promptable segmentation for any object, while ``segment everything" solves the downstream task of mask proposal generation for all objects. Since ``segment everything" does not necessarily require a prompt, FastSAM directly generates the mask proposal with YOLO v8 in a prompt-free manner. To enable promptable segmentation, a mapping algorithm is designed to select the mask from the proposal mask sets. It is worth highlighting that the follow-up works that evaluate its generalization/robustness or investigate its versatility mainly focus on the anything instead of everything mode because the former addresses the foundation task. Therefore, the comparison with FastSAM mainly focuses on ``segment anything", but we also provide a comparison regarding ``segment everything" for completeness.

\begin{wraptable}{r}{7cm}
\centering
\vspace{-5mm}
\caption{Comparison between FastSAM and MobileSAM.}
\label{tab:fastsam_vs_mobilesam}
\begin{tabular}{ccccc}
\toprule
 & FastSAM & MobileSAM  & Ratio\\ 
\midrule
Size & 68M & 9.66M & $ \approx$ 7 \\
Speed & 64ms & 12ms & $\approx$ 5 \\
\bottomrule
\end{tabular}
\end{wraptable}
\paragraph{MobileSAM is faster and smaller.} FastSAM consists of a YOLOv8-based detection branch and a YOLACT-based segmentation branch to perform a prompt-free mask proposal generation. It has 68M parameters and takes 40ms to process an image. By contrast, MobileSAM has less 10M parameters, which is significantly smaller. For the inference speed, on a single GPU, it takes 40ms to process an image while ours only takes 10ms, which is 4 times faster than FastSAM (see Table~\ref{tab:fastsam_vs_mobilesam}).

\begin{wraptable}{r}{7cm}
\centering
\vspace{-5mm}
\caption{mIoU comparison. With the assumption that the predicted mask from the original SAM is ground-truth, a higher mIoU indicates a better performance.}
\label{tab:mIoU_comparision}
\begin{tabular}{cccccc}
\toprule
   & 100 & 200 & 300  & 400 & 500 \\ 
\midrule
FastSAM  & 0.27 & 0.33 & 0.37 & 0.41 & 0.41 \\
MobileSAM  & 0.73 & 0.71 & 0.74 & 0.73 & 0.73 \\
\bottomrule
\end{tabular}
\end{wraptable}
\paragraph{mIoU comparison under segment anything mode.} We further compare the mIoU between the predicted masks with that of the original SAM. FastSAM is suggested to predict the mask with multiple points, for which we choose one for the foreground and the other for the background. The results in Table~\ref{tab:mIoU_comparision} show the mIoU for FastSAM is much smaller than that for MobileSAM, suggesting that the mask prediction of FastSAM is very different from that of the original SAM. Moreover, the mIoU for the FastSAM decreases very fast when the distance between the two prompt points. This is mainly caused by the fact that FastSAM often fails to predict the object when the foreground prompt point is set too close to the background prompt point.

\begin{figure*}[!htbp]
     \centering        
    \begin{minipage}[t]{0.9\textwidth}
         \includegraphics[width=\textwidth]{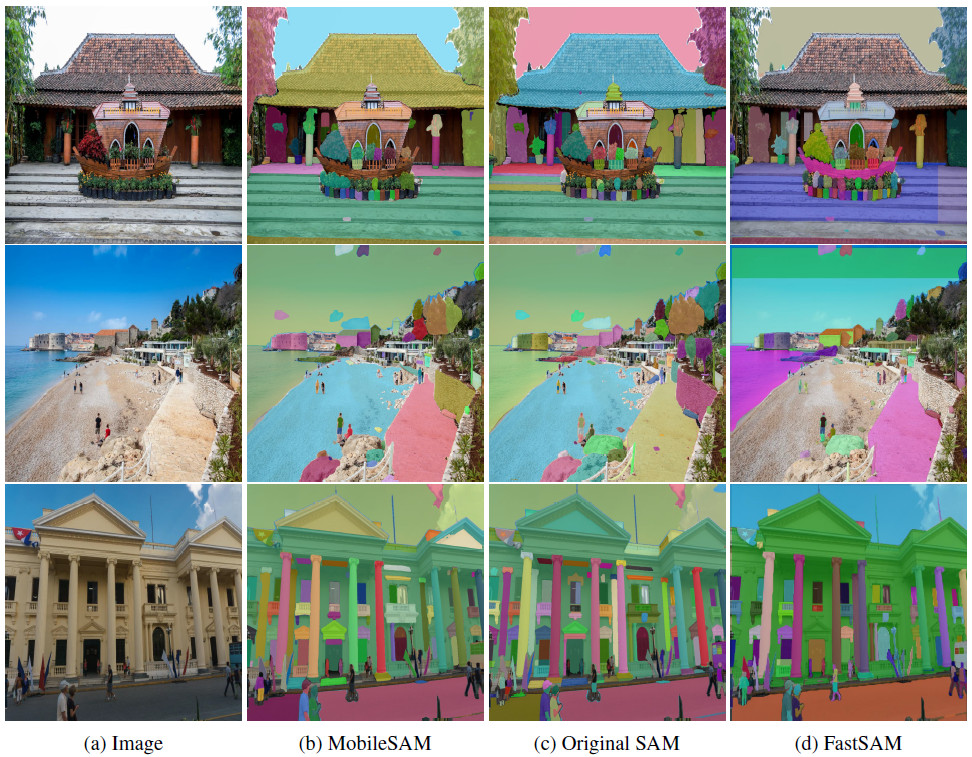}
     \end{minipage}
        \caption{Comparison of segment everything results.}
    \label{fig:everything}
\end{figure*}

\paragraph{Results for segment everything.} The results for ``segment everything" are shown in Figure~\ref{fig:everything}. For completeness, we also report the results of the original SAM, which generates a pleasing object proposal. We have two major observations. First, the results of our MobileSAM align surprisingly well with that of the original SAM. By contrast, the results of FastSAM are often less satisfactory. For example, FastSAM often fails to predict some objects, like the roof in the first image. Moreover, the mask proposal is sometimes difficult to interpret (see the mask for the stage in the first image and that for the sky in the second image). Second, FastSAM often generates masks that have non-smooth boundaries, for which we suggest the reader zoom in to check the details in Figure~\ref{fig:everything}. For example, the pillars in the third image have non-smooth boundaries, while the original SAM and our MobileSAM do not have this issue.

\section{Conclusion}
In this work, we aim to make SAM mobile-friendly by replacing the heavyweight image encoder with a lightweight one. We find that the naive way to train such a new SAM as in the original SAM paper leads to unsatisfactory performance, especially under a setup of limited training sources. The coupled optimization of the image encoder and mask decoder is the reason, and thus we propose decoupled distillation, whhere the knowledge is distilled from the image encoder ViT-H in the original SAM to a lightweight image encoder. We show that the resulting lightweight image encoder can be automatically compatible with the mask decoder in the original SAM. Our MobileSAM is more than 60 times smaller yet performs on par with the original SAM. Moreover, we conduct a comparison with the concurrent FastSAM and show that MobileSAM achieve superior performance. Our MobileSAM is also 4 times faster and 7 times smaller than the concurrent FastSAM, making it more suitable for mobile applications. Since our MobileSAM keeps all the pipeline of the original SAM and just replaces the image encoder, it can be plug-and-play for the existing SAM-based projects to move from a heavyweight SAM to a lightweight one with almost zero effort.

\bibliographystyle{unsrtnat}
\bibliography{bib_mixed,bib_local,bib_sam}

\end{document}